\documentclass[11pt]{article}

\usepackage[preprint]{acl}

\usepackage{times}
\usepackage{latexsym}
\usepackage{microtype}
\usepackage{hyperref}
\usepackage{url}
\usepackage{float}
\usepackage{amsmath}
\usepackage{booktabs}
\usepackage{tcolorbox}
\tcbuselibrary{listings}
\usepackage{multirow}
\usepackage{graphicx}
\usepackage{adjustbox}
\usepackage{amssymb}

\usepackage{lineno}
\usepackage{amsmath}

\definecolor{darkblue}{rgb}{0, 0, 0.5}
\hypersetup{colorlinks=true, citecolor=darkblue, linkcolor=darkblue, urlcolor=darkblue}

\usepackage[T1]{fontenc}

\usepackage[utf8]{inputenc}

\usepackage{microtype}

\usepackage{inconsolata}

\usepackage{graphicx}

\title{SAAG: Structured Agent Assessment and Grounding}

\author{
 \textbf{Ritvik Garimella \textsuperscript{1}},
 \textbf{Vedant Khandelwal \textsuperscript{1}},
 \textbf{Anvi Kohli \textsuperscript{1}},
 \textbf{Amit Sheth \textsuperscript{1,2}}
\\
\\
 \textsuperscript{1}Artificial Intelligence Institute, University of South Carolina \\
 \textsuperscript{2}Indian AI Research Organization
\\
 \small{
   \textbf{Correspondence:} \href{mailto:ritvikg@sc.edu}{ritvikg@sc.edu}
 }
}

\begin{document}
\maketitle
\begin{abstract}
Exact-match evaluation of agent-calling obscures qualitatively different failure modes: a model may select the right function yet hallucinate argument values, or satisfy a schema while choosing a agent for the wrong reason. Existing benchmarks collapse these distinctions into a single binary score, leaving practitioners unable to diagnose \emph{where} agent calls fail. We propose \textbf{SAAG} a cascaded diagnostic framework that decomposes agent-calling evaluation into three sequential stages: registry conformance, structural completeness, and argument grounding, each producing interpretable stage-specific diagnostics. These diagnostics additionally enable iterative self-repair: on prediction failure, the stage-specific signal guides targeted correction without leaking ground-truth values. We evaluate this framework on a controlled benchmark derived from Glaive's function-calling dataset across registry sizes of 5, 10, and 15 agents using three local sub-4B-parameter models. Structured feedback consistently improves argument precision and reduces value hallucination relative to single-pass inference and uninformative binary feedback, while end-to-end F1 gains are modest and model-dependent. These results suggest that stage-decomposed diagnostic evaluation is a necessary lens for understanding and improving agent-calling reliability across model families and registry scales.
\end{abstract}

\section{Introduction}

Function calling has become the operational interface for LLMs to interact with APIs, search systems, and external software. The central evaluation question is no longer whether a model can emit valid JSON, but whether it can choose the \emph{right} tool for the \emph{right} reason. A model may produce syntactically correct calls yet fail in deployment by selecting a tool whose capability does not match the request, or by supplying arguments that are not grounded in the user's actual query. Such failures are silent but costly: the call passes format checks while remaining operationally wrong.

Recent works: API-Bank, ToolLLM, Gorilla, APIBench, and BFCL, has established strong baselines for tool use and API invocation \citep{li2023apibank,qin2024toolllm,patil2024gorilla,patil2025bfcl}. Yet the dominant evaluation paradigm still centres on \emph{formal correctness}: whether the model emitted a registered function name, whether calls are executable, or whether execution succeeds end-to-end. These metrics do not answer the more critical question: \emph{did the model select the tool because it correctly understood its capabilities?} Existing evaluations are strong at measuring conformance but weaker at isolating \emph{where and why} a tool call fails. Recent work on tool hallucination confirms this concern: Reliability Alignment distinguishes tool selection hallucination from tool usage hallucination \citep{xu2025reducing}, and ToolBeHonest shows that even frontier models struggle when tool availability and task solvability are systematically varied \citep{zhang2024toolbehonest}. Valid tool calls are not sufficient evidence of grounded tool understanding.

A second limitation lies in evaluation granularity. A call can fail because the model hallucinated a function name, omitted a required argument, introduced a spurious parameter, or invented an unsupported argument value. Existing benchmarks collapse these qualitatively different failure modes into a single score, leaving practitioners unable to identify whether a model failed at function selection, schema compliance, or argument grounding. This conflation obscures both diagnosis and targeted improvement. 


We propose \textsc{SAAG} (\textbf{S}tructured \textbf{A}gent \textbf{A}ssessment and \textbf{G}rounding), a cascaded diagnostic framework that decomposes tool calling into three sequential stages: \emph{registry conformance} (is the function name in the registry?), \emph{structural completeness} (do the arguments satisfy the schema?), and \emph{argument grounding} (are argument values supported by the query?). Each stage produces interpretable, stage-specific metrics that can be applied to any tool-calling system independently of any correction strategy. As one instantiation of these diagnostics, we additionally implement a cascaded correction loop that returns minimal structured feedback on failure without leaking ground truth. This toolkit acts as a sub-module of the larger multi-agent orchestration framework DYNO \citep{garimella2026dyno} with evaluations scheme inspired from conceptualization in C3AN \cite{sheth2025composite}.


\paragraph{Scope.} Our study focuses on \emph{single-tool selection from a fixed registry}: a controlled yet important step, setting the foundation for multi-agent orchestration. We evaluate three sub-4B models across registry sizes of 5, 10, and 15 agents, comparing three feedback regimes: direct inference, binary-feedback iteration, and full \textsc{SAAG} structured feedback. We do not claim to address multi-agent orchestration, dynamic agent discovery, or real agent execution; our goal is to isolate the diagnostic question: \textit{when agent registries become confusable, can stage-decomposed diagnostic metrics expose where and why agent calls fail, and can structured feedback act on those diagnostics to improve reliability?}

\paragraph{Contributions}

\begin{enumerate}
    \item We introduce \textsc{SAAG}'s eight intrinsic grounding metrics that act as a diagnostic measure that 
    characterizes agent-calling failures at a granular level.
    \item We demonstrate an application of the diagnostics as a cascaded 
    correction loop across three local 
    small model families.
\end{enumerate}

\begin{figure*}[t]
\centering
\includegraphics[width=0.95\textwidth]{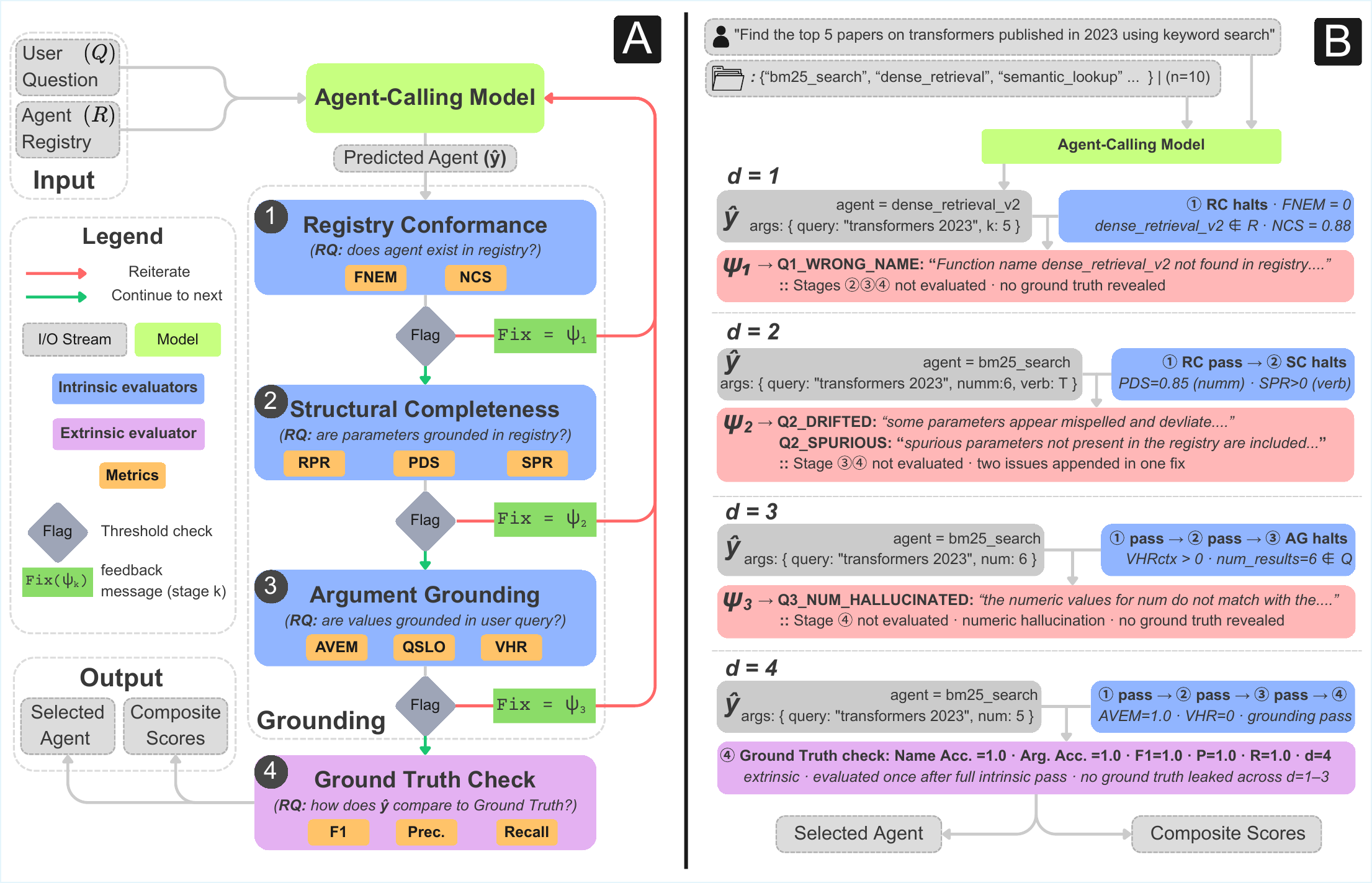}
\caption{\textbf{The SAAG framework.} (A) Overview of the cascaded evaluation pipeline: a tool-calling model's prediction passes sequentially through Registry Conformance (Stage~1), Structural Completeness (Stage~2), and Argument Grounding (Stage~3), each producing interpretable metrics and a targeted correction signal $\psi_k$ on failure. Ground truth (\textit{extrinsic} evaluation) is assessed only after all stages pass. (B) A worked example illustrating four correction iterations ($d=1$--$4$): the model first hallucinates a function name (FNEM=0), then produces drifted and spurious parameters, then introduces a numeric hallucination, before arriving at a fully grounded, schema-conformant call at $d=4$.}
\label{fig:main}
\end{figure*}

\section{Structured Agent Assessment and Grounding}
\label{sec:methodology}

SAAG decomposes tool-calling evaluation into three sequential quality gates: Registry Conformance (\S\ref{sec:q1}), Structural Completeness (\S\ref{sec:q2}), and Argument Grounding (\S\ref{sec:q3}). The cascade is strict: evaluation halts at the first failing stage, ensuring that downstream metrics are computed only on predictions that satisfy upstream obligations. A structured correction $\psi_k$ is returned to the model at each failure, enabling iterative self-repair without leaking ground-truth values. An important consequence of this design is that intrinsic metrics from later stages are \emph{stage-conditional diagnostics}: they characterise the quality of predictions that already passed earlier gates, and should not be interpreted as unconditional population statistics. We return to this point when reporting results.

\subsection{Registry Conformance}
\label{sec:q1}

The first obligation is that the predicted function name $\hat{y}_\text{name}$ corresponds to a registered tool. Let $\mathcal{R}$ denote the set of all function names in the system prompt. We define the \textit{Function Name Exact Match} (FNEM) as:

\begin{equation}
    FNEM (\hat{y}_\text{name},\, \mathcal{R})=
    \begin{cases}
        -1 & \text{if } \hat{y}_\text{name} =  \varnothing \\
        0 & \text{if } \hat{y}_\text{name} \notin \mathcal{R} \\
        1 & \text{if } \hat{y}_\text{name} \in \mathcal{R}
    \end{cases}
\end{equation}

The three cases distinguish complete parse failure ($-1$), registry mismatch ($0$), and conformance ($1$). When FNEM $= 0$, we compute a \textit{Nearest Candidate Score} (NCS) to quantify drift from the closest valid name:

\begin{equation}
\begin{aligned}
\text{NCS}(\hat{y}_\text{name}, \mathcal{R})
&= \max_{c \in \mathcal{C}} \Big[
0.6\,\text{WRatio}(\hat{y}_\text{name}, c) \\
&\quad + 0.4\,\text{Jaccard}(\hat{y}_\text{name}, c)
\Big].
\end{aligned}
\end{equation}


where $\mathcal{C} \subseteq \mathcal{R}$ are candidates sharing at least one trigram with $\hat{y}_\text{name}$, WRatio is the weighted partial ratio from RapidFuzz \citep{rapidfuzz} (normalised to $[0,1]$), and Jaccard is computed over whitespace-tokenised lowercased strings. The reason to use $0.6$ as weightage for WRatio as it better captures approximate string similarity under minor variations in comparison to Jaccard that provides complementary token-level overlap. \textbf{What this catches:} hallucinated or misspelled function names. A high NCS with FNEM $= 0$ indicates a recoverable misspelling; a low NCS indicates fabrication. \textbf{What this cannot capture:} whether the \emph{correct} registry name actually corresponds to the tool best suited for the query. If FNEM $= 1$, Stage~1 passes; otherwise the correction $\psi_1$ identifies whether the name was absent or unrecognised and embeds the nearest registry match to guide revision without revealing ground truth. Fix templates are provided in Appendix~\ref{app:fixes}.

\subsection{Structural Completeness}
\label{sec:q2}

Given a conformant function name, the second obligation is that the predicted argument set $P$ satisfies the function's schema. Let $A_\text{req}$ and $A_\text{all}$ denote the required and admissible parameter sets respectively. We measure structural completeness along three axes.

\paragraph{Required Parameter Recall (RPR)} measures coverage of mandatory arguments:
\begin{equation}
    \text{RPR} = \frac{|A_\text{req} \cap P|}{|A_\text{req}|}
\end{equation}
RPR $< 1$ is a hard failure: missing required parameters will cause execution to fail regardless of value correctness.

\paragraph{Parameter Drift Score (PDS)} captures whether out-of-schema predictions are recoverable misspellings rather than hallucinated arguments. For each parameter $p \in P \setminus A_\text{all}$:
\begin{equation}
    \text{PDS}(p,\, A_\text{all}) = \max_{a \in A_\text{all}}\; \text{WRatio}(p,\, a) \;/\; 100
\end{equation}
A parameter is classified as \textit{drifted} if $\text{PDS} \geq 0.8$ and its nearest match is not already present as an exact prediction, and as \textit{spurious} otherwise. Drifted parameters receive a targeted spelling fix; spurious ones are flagged for removal.

\paragraph{Spurious Parameter Rate (SPR)} measures the proportion of predicted arguments with no recoverable schema grounding:
\begin{equation}
    \text{SPR} = \frac{\big|\{p \in P \setminus A_\text{all} : \text{PDS}(p, A_\text{all}) < 0.8\}\big|}{|P|}
\end{equation}

\paragraph{Stage 2 pass criterion.} The stage passes if and only if RPR $= 1$ and SPR $= 0$. Separately, drifted parameters ($\text{PDS} \geq 0.8$) always trigger a targeted correction regardless of the formal pass/fail outcome, because unresolved drift propagates into downstream grounding failures even when the required parameters are all present. The correction $\psi_2$ is stratified by failure type: missing required parameters, spurious inclusions, and drifted spellings each produce distinct guidance. \textbf{What this catches:} schema violations, missing, misspelled, and fabricated parameters. \textbf{What this cannot capture:} whether \emph{correct} parameter names carry appropriate values.

\subsection{Argument Value Grounding}
\label{sec:q3}

The third obligation is that argument values are faithfully grounded in the user query $Q$. Unlike the previous stages, grounding is continuous: a model may produce a paraphrase, a near-miss, or an outright hallucination.

\paragraph{Argument Value Exact Match (AVEM)} measures the rate at which predicted values appear verbatim in $Q$:
\begin{equation}
    \text{AVEM} = \frac{\big|\{i : \text{ExactMatch}(v_i, Q) = 1\}\big|}{|P|}
\end{equation}
The check is type-aware: string values are matched case-insensitively at word boundaries; numeric values must match a floating-point token in $Q$; list values require all elements to satisfy the condition individually. Exact match is the \emph{only} grounding check applied to numeric arguments, there is no partial-credit fallback.

\paragraph{Query-Supported Lexical Overlap (QSLO)} measures lexical proximity exclusively for string arguments that fail exact match. Let $\mathcal{S} = \{i : \text{type}(v_i) = \text{string} \wedge \text{ExactMatch}(v_i, Q) = 0\}$. Then:
\begin{equation}
    \text{QSLO} = \frac{1}{|\mathcal{S}|} \sum_{i \in \mathcal{S}} \text{PartialRatio}(v_i,\, Q) \;/\; 100
\end{equation}
where PartialRatio scores the best contiguous alignment of $v_i$ within $Q$ via RapidFuzz~\citep{rapidfuzz}. This is a \emph{lexical} overlap measure, not a semantic similarity score; it captures whether argument values are extractable substrings of the query, not whether they are semantically equivalent. \textbf{What this catches:} paraphrased or truncated query spans. \textbf{What this cannot capture:} semantically valid values that use entirely different surface forms than the query.

\paragraph{Value Hallucination Rate (VHR)} aggregates per-argument hallucination:

\begin{equation}
h(v_i,Q)=
\begin{cases}
0, &  \mathrm{EM}(v_i,Q)=1,\\
1-\frac{\mathrm{PR}(v_i,Q)}{100}, & \mathrm{str}(v_i),\\
1, & \mathrm{num}(v_i)\wedge \mathrm{HN}(Q)\\
\text{excluded}, & \text{otherwise.}
\end{cases}
\end{equation}

\noindent
Here, $\mathrm{EM}$ denotes \textit{ExactMatch}, $\mathrm{PR}$ denotes 
\textit{PartialRatio}, and $\mathrm{HN}$ denotes \textit{HasNumbers}. 
We use $\mathrm{str}(\cdot)$ and $\mathrm{num}(\cdot)$ to indicate string 
and numeric types, respectively.

The final case covers arguments for which no grounding context exists in the query (e.g., a ticker symbol inferred from a company name). These are treated as \textit{assumed} and excluded from the pass/fail decision, since penalising them would conflate hallucination with legitimate model inference. The pass criterion is defined over the contextual subset $\mathcal{G} = \{i : \text{type}(v_i) = \text{string}\} \cup \{i : \text{type}(v_i) = \text{numeric} \wedge \text{HasNumbers}(Q)\}$:
\begin{equation}
    \text{VHR}_\text{ctx} = \sum_{i \in \mathcal{G}} h(v_i, Q) = 0
\end{equation}
A non-zero $\text{VHR}_\text{ctx}$ produces the correction $\psi_3$, stratified by failure type: numeric mismatches, hard-drifted strings ($\text{PartialRatio} < 0.6$), and soft-drifted strings each receive distinct correction language.

\subsection{Cascaded Correction Loop}
\label{sec:loop}

The three stages are evaluated sequentially. The cascade halts at the first failing stage $k$, and its structured fix $\psi_k$ is appended to the conversation as a user turn:
\begin{equation}
    \hat{y}^{(t+1)} = \mathcal{M}\!\left(H^{(t)} \circ \psi_k\right)
\end{equation}
where $H^{(d)}$ is the conversation history at iteration $d$. The loop repeats up to $D_\text{max} = 15$ iterations and terminates early on a full pass. Each fix $\psi_k$ is \emph{templated and deterministic}: given the same failure mode and metric values, the same correction string is produced, ensuring reproducibility. The full set of fix templates is given in Appendix~\ref{app:fixes}.

\section{Experimental Setup}
\label{sec:experiments}

We evaluate whether cascaded diagnostic feedback improves tool-calling reliability in sub-4B-parameter models when (a) model families change, (b) registry scale increases, and (c) feedback specificity varies.

\paragraph{Dataset construction.} We construct a controlled benchmark from the Glaive function-calling dataset~\citep{glaiveai2023functioncalling}. From the original corpus, we retain only rows that require exactly one tool call and for which the system prompt can be standardised to expose functions as JSON schema blocks with typed parameters and explicit required fields. Rows with malformed schemas, multi-call sequences, or ambiguous ground truth are excluded. After filtering, 3{,}000 queries form the evaluation set. For each query, registry scale is controlled by retaining the ground-truth function and sampling $S-1$ distractors \emph{uniformly at random} from the remaining function pool to reach $S \in \{5, 10, 15\}$ functions per prompt. Distractors are not curated for semantic confusability; they are random draws, meaning that registry difficulty comes from scale, not from adversarial similarity. The same 3{,}000 queries are reused across all registry scales so that performance differences reflect registry complexity alone.

\paragraph{Models.} We evaluate three models: Llama~3.2:3B~\citep{grattafiori2024llama}, Ministral~3B~\citep{liu2026ministral}, and Granite~3.1:3B~\citep{mishra2024granite}. All inferences are run locally via Ollama at temperature $\tau = 0$, with no fine-tuning; every condition is a pure inference-time intervention. This selection spans three distinct families, testing whether SAAG's feedback generalises across architectural and training differences.

\paragraph{Feedback modes.} Three conditions are compared (Table~\ref{tab:conditions}):
\textbf{Direct pass}: a single inference pass with no correction loop, providing a zero-feedback baseline of out-of-the-box capability.
\textbf{Binary feedback}: the correction loop is active, but each stage failure returns a fixed uninformative signal (\texttt{``No''}) without diagnostic content. This serves as an \emph{ablation} to isolate the effect of iterative re-prompting from the information content of the feedback; it is not intended as a representative real-world baseline.
\textbf{Structured feedback (SAAG)}: the full cascaded diagnostic framework from \S\ref{sec:methodology}, where each failing stage returns a stage-stratified correction $\psi_k$ targeted to the specific failure mode.
Direct pass uses a single attempt ($D = 1$). Binary feedback and SAAG both operate under a maximum of $D_\text{max} = 15$ iterations, bounding cost fairly across loop-enabled conditions.

\paragraph{Definitions.}
\textit{Solve rate} is the fraction of the 3{,}000 queries for which the model produces a prediction that passes all three SAAG stages and matches the ground-truth function name \emph{within the iteration budget}. For direct pass, this is identical to single-pass accuracy; for loop-enabled modes, it captures whether the loop eventually recovers a correct call.

\paragraph{Metric classes.} We report two classes of metrics. \textit{Intrinsic metrics} are the eight SAAG diagnostics from \S\ref{sec:methodology}, aggregated across rows to characterise grounding quality per stage. Because the cascade is strict, intrinsic metrics from later stages are conditioned on passing earlier stages and should be interpreted as stage-conditional diagnostics, not unconditional population statistics. \textit{Extrinsic metrics} (name accuracy, argument precision, recall, F1) are computed against ground truth for all rows and serve as end-to-end validity checks.

\paragraph{Uncertainty.} All $\pm$ values in Tables~\ref{tab:extrinsic}--\ref{tab:intrinsic} are standard deviations across the three registry scales ($S \in \{5, 10, 15\}$), reflecting variability due to registry complexity rather than sampling noise.

\begin{table}[h]
\centering
\small
\renewcommand{\arraystretch}{1.4}
\begin{tabular}{llcc}
\toprule
\textbf{Mode} & \textbf{CS} & \textbf{Loop} & \textbf{$D_\text{max}$} \\
\midrule
Direct pass   & None                        & No  & 1             \\
Binary feedback  & Fixed \texttt{``No''}    & Yes & 15     \\
SAAG          & targeted diagnostic & Yes & 15     \\
\bottomrule
\end{tabular}
\caption{Experimental conditions. All three modes are evaluated across three models and three registry scales ($S \in \{5, 10, 15\}$), yielding 27 conditions on 3{,}000 queries each. Binary feedback is an ablation isolating the effect of iteration from feedback content; it is not a strong external baseline. CS - Correction Signal}
\label{tab:conditions}
\end{table}

\section{Results}
\label{sec:results}

We report intrinsic grounding diagnostics (Table~\ref{tab:intrinsic}) and extrinsic correctness metrics (Table~\ref{tab:extrinsic}), both averaged across registry scales $S \in \{5, 10, 15\}$. Intrinsic metrics are the primary lens; extrinsic results serve as an end-to-end validity check. Per-scale breakdowns are in Appendix~\ref{app:res}.

\paragraph{Argument grounding is the dominant failure mode.}
Across all models and conditions, name accuracy is near ceiling while argument accuracy lags substantially. This persistent gap confirms argument grounding as the primary bottleneck. Conflating these failure modes into a single outcome metric, as standard exact-match benchmarks do, obscures where models most need improvement.


\paragraph{Precision improves consistently; F1 gains are mixed.}
In agent-call grounding, a hallucinated argument value causes silent downstream failures, whereas a missing argument is at least detectable. Precision is therefore the more meaningful signal in this setting. SAAG improves argument precision for all three models and solve rate improves substantially across all models (Table~\ref{tab:extrinsic}), with the largest gain for LLaMA ($0.615 \to 0.802$). End-to-end F1, however, tells a more nuanced story: SAAG improves F1 for LLaMA ($0.527 \to 0.533$) but slightly decreases it for Ministral and Granite, driven by recall loss. This is consistent with an overcorrection effect in reasoning models: both Ministral~3 and Granite~4 are reasoning-oriented models, and iterative feedback loops can cause them to revise initially correct predictions, a behaviour documented for Ministral~3~\cite{liu2026ministral} and similarly observed for Granite~4.

\paragraph{Structured feedback consistently reduces hallucination.}
The most reliable gain from structured feedback (SAAG) is reduced argument value hallucination. VHR decreases under SAAG for every model: from $0.336$ to $0.311$ for LLaMA, from $0.277$ to $0.166$ for Ministral (a 40\% reduction), and from $0.278$ to $0.259$ for Granite. AVEM improves correspondingly, with the largest gain for Ministral ($0.658 \to 0.786$). Binary feedback, by contrast, occasionally \textit{increases} VHR (Granite: $0.278 \to 0.288$), suggesting that for reasoning enabled models, correction signals introduce spurious drift rather than fix grounding errors.

\paragraph{SAAG converges faster than binary feedback.}
Structured feedback converges within 2-3 iterations across all registry scales (Figure~\ref{fig:cumulative}), reaching a higher ceiling than binary feedback at every depth i.e., consuming less overall time. This confirms that \emph{content} based diagnostic correction drives grounding recovery quicker than pure iterative looping.

\begin{table*}[t]
\centering
\setlength{\tabcolsep}{3pt}
\renewcommand{\arraystretch}{1.2}
\scriptsize
\begin{tabular}{@{}l ccc ccc ccc@{}}
\toprule
& \multicolumn{3}{c}{\textbf{LLaMA 3.2}}
& \multicolumn{3}{c}{\textbf{Ministral}}
& \multicolumn{3}{c}{\textbf{Granite}} \\
\cmidrule(lr){2-4} \cmidrule(lr){5-7} \cmidrule(lr){8-10}
\textbf{Metric} & Direct & Binary & SAAG
                & Direct & Binary & SAAG
                & Direct & Binary & SAAG \\
\midrule
Precision
    & 0.524{\tiny$\pm$.037} & 0.525{\tiny$\pm$.035} & 0.540{\tiny$\pm$.033}
    & 0.705{\tiny$\pm$.013} & 0.648{\tiny$\pm$.013} & 0.661{\tiny$\pm$.013}
    & 0.747{\tiny$\pm$.015} & 0.667{\tiny$\pm$.013} & 0.717{\tiny$\pm$.049} \\
Recall
    & 0.537{\tiny$\pm$.037} & 0.524{\tiny$\pm$.033} & 0.534{\tiny$\pm$.031}
    & 0.710{\tiny$\pm$.013} & 0.648{\tiny$\pm$.013} & 0.654{\tiny$\pm$.013}
    & 0.745{\tiny$\pm$.015} & 0.663{\tiny$\pm$.013} & 0.690{\tiny$\pm$.032} \\
F1
    & 0.527{\tiny$\pm$.370} & 0.521{\tiny$\pm$.033} & 0.5330{\tiny$\pm$.032}
    & 0.705{\tiny$\pm$.013} & 0.646{\tiny$\pm$.013} & 0.654{\tiny$\pm$.013}
    & 0.744{\tiny$\pm$.015} & 0.664{\tiny$\pm$.013} & 0.698{\tiny$\pm$.038} \\

Solve Rate 
& 0.615{\tiny$\pm$.016} & 0.768{\tiny$\pm$.014} & 0.802{\tiny$\pm$.014}
    & 0.726{\tiny$\pm$.008} & 0.829{\tiny$\pm$.000} & 0.825{\tiny$\pm$.005}
    & 0.694{\tiny$\pm$.020} & 0.816{\tiny$\pm$.002} & 0.771{\tiny$\pm$.048} \\
\bottomrule
\end{tabular}
\caption{Extrinsic metrics averaged across registry scales $S \in \{5,10,15\}$; $\pm$ values are standard deviations across scales. Higher is better for all metrics ($\uparrow$). SAAG improves precision for all models but F1 gains are mixed: LLaMA improves, while Ministral and Granite show slight decreases due to recall loss from overcorrection. Per-scale breakdowns in Appendix~\ref{app:res}.}
\label{tab:extrinsic}
\end{table*}

\begin{figure*}[t]
    \centering
    \includegraphics[width=\textwidth]{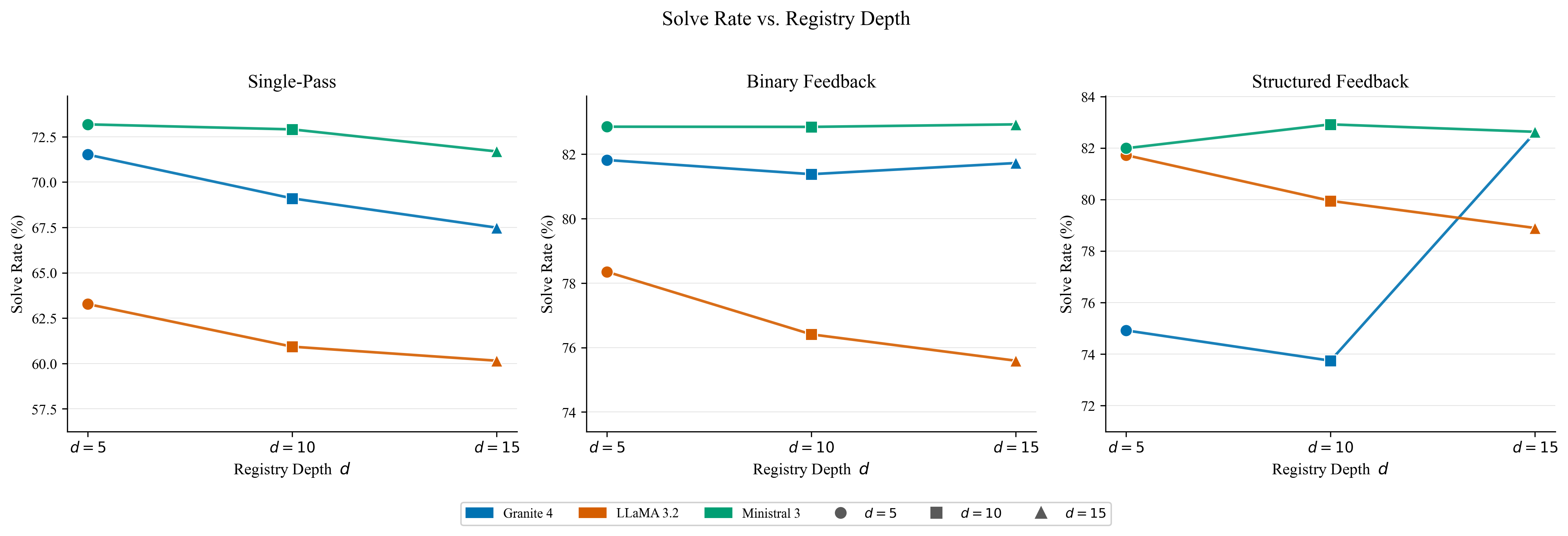}
    \caption{Solve rate (\%) across registry depths (d = 5, 10, 15) for three models under Single-Pass, Binary Feedback, and Structured Feedback conditions. Ministral 3 consistently leads across all settings, while LLaMA 3.2 degrades most sharply with increasing depth. Structured Feedback provides the greatest gains and most stable performance as registry size scales.}
    \label{fig:solve_rate}
\end{figure*}

\begin{figure*}[t]
    \centering
    \includegraphics[width=\textwidth]{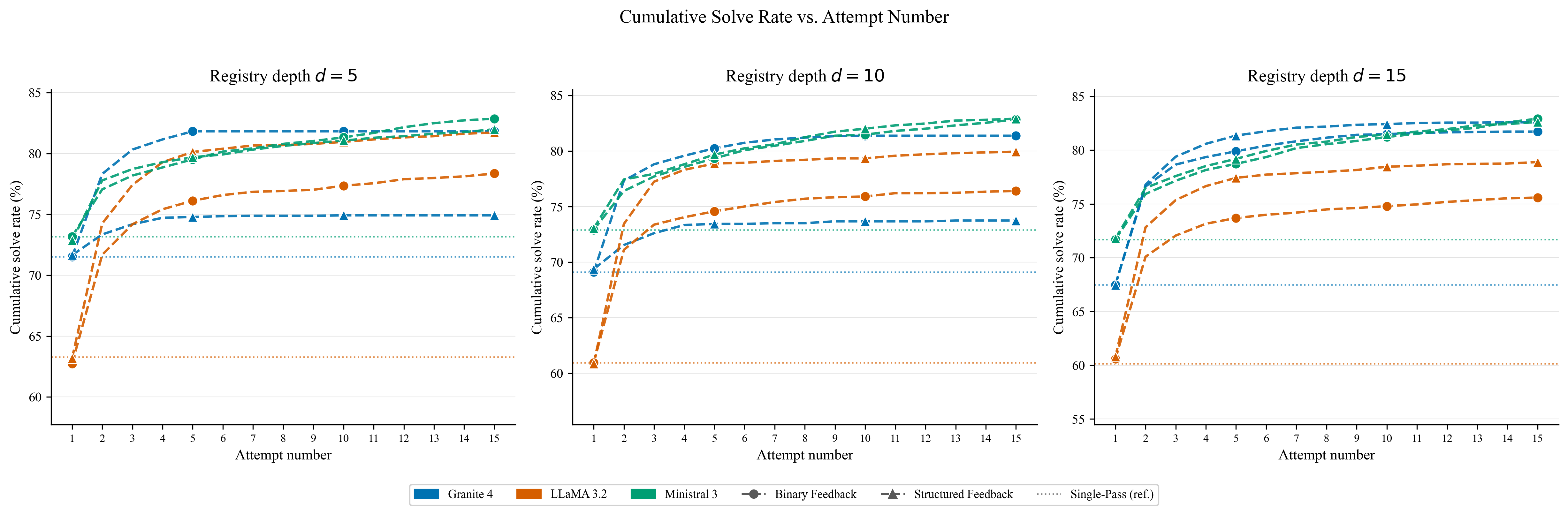}
    \caption{Cumulative solve rate (\%) over up to 15 attempts across registry depths (d = 5, 10, 15) for Granite 4, LLaMA 3.2, and Ministral 3 under Binary and Structured Feedback, with Single-Pass as the dotted baseline. All models converge within the first 2–3 attempts, with Structured Feedback improving more with attempts than Binary Feedback across all depths. Lower registry depths converge faster compared to higher depths indicating that more complexity in registry requires more attempts to solve.}
    \label{fig:cumulative}
\end{figure*}

\begin{figure*}[t]
    \centering
    \includegraphics[width=\textwidth]{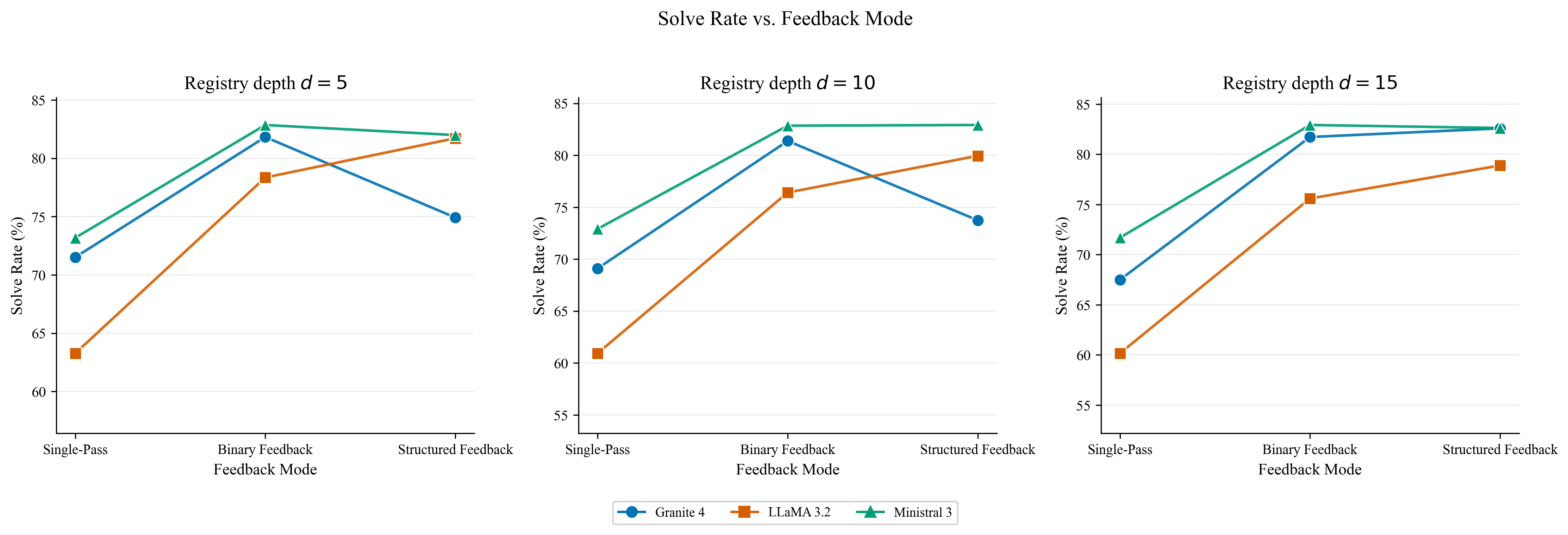}
    \caption{Solve rate (\%) across feedback modes (Single-Pass, Binary, Structured) at registry depths d = 5, 10, 15 for Granite 4, LLaMA 3.2, and Ministral 3. All models improve substantially from Single-Pass to Binary Feedback, with LLaMA 3.2 showing the largest gains. Ministral 3 remains stable from Binary to Structured Feedback, while Granite 4 drops notably at d = 5, suggesting it is more sensitive to feedback at shallower depths due to reasoning enabled.}
    \label{fig:cumulative}
\end{figure*}

\begin{table*}[h!]
\centering
\setlength{\tabcolsep}{2.5pt}
\renewcommand{\arraystretch}{1.2}
\scriptsize
\begin{tabular}{@{}ll ccc ccc ccc@{}}
\toprule
& & \multicolumn{3}{c}{\textbf{LLaMA 3.2}}
& \multicolumn{3}{c}{\textbf{Ministral}}
& \multicolumn{3}{c}{\textbf{Granite}} \\
\cmidrule(lr){3-5} \cmidrule(lr){6-8} \cmidrule(lr){9-11}
\textbf{Stage} & \textbf{Metric} & Direct & Binary & SAAG
                                 & Direct & Binary & SAAG
                                 & Direct & Binary & SAAG \\
\midrule
\multirow{2}{*}{\textsc{RC}}
& FNEM $\uparrow$
    & 0.999{\tiny$\pm$.000} & 0.9977{\tiny$\pm$.001} & 1.0000{\tiny$\pm$.000}
    & 1.000{\tiny$\pm$.000} & 1.000{\tiny$\pm$.000} & 0.999{\tiny$\pm$.001}
    & 1.000{\tiny$\pm$.000} & 0.998{\tiny$\pm$.002} & 0.998{\tiny$\pm$.002} \\
\midrule
\multirow{2}{*}{\textsc{SC}}
& RPR $\uparrow$
    & 0.968{\tiny$\pm$.012} & 0.965{\tiny$\pm$.004} & 0.963{\tiny$\pm$.017}
    & 0.984{\tiny$\pm$.005} & 0.937{\tiny$\pm$.002} & 0.886{\tiny$\pm$.008}
    & 0.958{\tiny$\pm$.021} & 0.976{\tiny$\pm$.012} & 0.926{\tiny$\pm$.035} \\
& SPR $\downarrow$
    & 0.043{\tiny$\pm$.028} & 0.015{\tiny$\pm$.006} & 0.028{\tiny$\pm$.039}
    & 0.008{\tiny$\pm$.006} & 0.001{\tiny$\pm$.001} & 0.002{\tiny$\pm$.001}
    & 0.031{\tiny$\pm$.015} & 0.014{\tiny$\pm$.008} & 0.020{\tiny$\pm$.007} \\
\midrule
\multirow{3}{*}{\textsc{AG}}
& AVEM $\uparrow$
    & 0.584{\tiny$\pm$.036} & 0.613{\tiny$\pm$.010} & 0.617{\tiny$\pm$.053}
    & 0.658{\tiny$\pm$.004} & 0.717{\tiny$\pm$.003} & 0.786{\tiny$\pm$.009}
    & 0.648{\tiny$\pm$.003} & 0.645{\tiny$\pm$.003} & 0.688{\tiny$\pm$.030} \\
& QSLO $\uparrow$
    & 0.339{\tiny$\pm$.019} & 0.347{\tiny$\pm$.006} & 0.343 {\tiny$\pm$.012}
    & 0.268{\tiny$\pm$.010} & 0.331{\tiny$\pm$.011} & 0.361{\tiny$\pm$.035}
    & 0.318{\tiny$\pm$.008} & 0.296{\tiny$\pm$.006} & 0.291{\tiny$\pm$.005} \\
& VHR $\downarrow$
    & 0.336{\tiny$\pm$.028} & 0.314{\tiny$\pm$.006} & 0.311 {\tiny$\pm$.041}
    & 0.277{\tiny$\pm$.006} & 0.213{\tiny$\pm$.004} & 0.166{\tiny$\pm$.008}
    & 0.278{\tiny$\pm$.005} & 0.288{\tiny$\pm$.002} & 0.259{\tiny$\pm$.021} \\
\bottomrule
\end{tabular}
\caption{Intrinsic grounding diagnostics averaged across registry scales $S \in \{5,10,15\}$; $\pm$ values are standard deviations across scales. These are \emph{stage-conditional} metrics: later-stage values are computed only on predictions that passed earlier stages. NCS and PDS are omitted from the main table as they are near-constant across conditions; full values including per-scale breakdowns are in Appendix~\ref{app:res}.}
\label{tab:intrinsic}
\end{table*}

\section{Related Work}

\paragraph{Tool-calling evaluation and benchmarks.}
Recent benchmarks such as API-Bank~\citep{li2023apibank}, ToolLLM~\citep{qin2024toolllm}, Gorilla~\citep{patil2024gorilla}, and BFCL~\citep{patil2025bfcl} have established standardized protocols for evaluating tool use in large language models. While effective for measuring overall performance, these benchmarks predominantly focus on format correctness or end-to-end execution success, often conflating distinct failure modes into a single aggregate metric. As a result, they provide limited diagnostic insight into where and why failures occur. ToolBeHonest~\citep{zhang2024toolbehonest} advances this direction by introducing a structured, multi-level diagnostic framework that decomposes tool use into solvability detection, solution planning, and missing-tool analysis, while also evaluating robustness under hallucination-inducing scenarios such as missing or constrained tools. However, its scope is limited to evaluation. In contrast, SAAG extends this paradigm beyond diagnosis by not only identifying stage-specific failure modes but also leveraging them to drive an inference-time, iterative self-repair process that actively corrects tool-calling errors.

\paragraph{Tool hallucination.}
Tool hallucination has emerged as a critical challenge in reliable tool use. Reliability Alignment~\citep{xu2025reducing} distinguishes between \emph{tool selection hallucination} and \emph{tool usage hallucination}, highlighting the need for stage-aware analysis of model behavior. Prior approaches address these issues through additional training signals, such as process-level reward models~\citep{qian2025toolrlrewardtoollearning} or self-reflection and feedback mechanisms~\citep{masterman2024landscapeemergingaiagent}. In contrast, SAAG operates entirely at inference time, leveraging deterministic and templated feedback without requiring auxiliary models or retraining, thereby offering a lightweight yet effective alternative.

\paragraph{Iterative correction.}
Frameworks such as ReAct~\citep{yao2022react} and related reasoning-and-acting paradigms improve tool use through iterative interaction, enabling models to refine their outputs over multiple steps. However, the feedback signals in these approaches are typically coarse and undifferentiated. In this work, we explicitly disentangle the role of iteration from the quality of feedback through a binary feedback ablation, and ground our correction signals in deterministic, rule-based templates derived from intrinsic stage metrics rather than 
a secondary LLM functioning as judge.

\section{Discussion}
\label{discussion}

\paragraph{Summary.} 
SAAG's stage-decomposed metrics reveal what aggregate benchmarks cannot: argument grounding is the dominant failure mode, overcorrection is a distinct and measurable phenomenon, and precision gains do not always propagate to F1. These findings hold independently of any correction strategy with the intrinsic diagnostics characterizing model behaviour whether or not the correction loop is applied. Where the loop (diagnostic feedback) is applied, the strongest signal is that \emph{telling a model what failed} is substantially more effective than \emph{telling it that it failed}, a finding that holds across all three model families and registry complexities. Furthermore, the stage-conditional design further enables actionable diagnosis. For example: under binary feedback, Granite's RPR improves while VHR worsens simultaneously; an indication that a single aggregate metric would simply average out and miss. Per-stage decomposition surfaces such contradictions directly and points to 
where intervention is needed.

\paragraph{Limitations.}
Several limitations bound the scope of our findings. Firs, our evaluation is restricted to single-agent selection from a fixed registry; multi-agent sequences, dynamic registries, and agent chaining introduce failure modes that SAAG does not currently address. Second, distractors are sampled uniformly at random rather than for semantic confusability, so our registry difficulty is a lower bound on what real deployments face. Third, the argument grounding stage relies on lexical overlap (PartialRatio) rather than true semantic similarity; values that are semantically correct but lexically distant from the query will be incorrectly flagged. Fourth, the correction loop can overcorrect valid predictions, particularly for agent names, suggesting that tighter thresholds or confidence-gated correction are needed. Also, these overcorrections occur at a higher frequency for reasoning models such as Ministral~3 and Granite~4.  Finally, we evaluate only sub-4B-parameter models at inference time with no fine-tuning; whether the diagnostic decomposition transfers to larger models or training-time objectives remains open.

\section{Conclusion}
We presented SAAG, a diagnostic framework that decomposes agent-calling 
evaluation into three sequential stages: registry conformance, structural 
completeness, and argument grounding, with each having interpretable, stage-specific 
metrics. These metrics constitute the primary contribution: \textit{they expose where agent calls fail i.e., at function selection, schema compliance, or argument grounding and at a granularity} that existing exact-match benchmarks do not provide.Evaluated across three sub-4B model families and three registry scales, the cascaded correction loop consistently improves argument precision and reduces value hallucination, while also surfacing a measurable overcorrection effect in reasoning models, name selection and recall,a limitation that motivates confidence-gated correction as a direction for future work. The central finding is that telling a model \textit{what} failed is substantially more 
effective than telling it \textit{that} it failed.

\bibliography{custom}

@inproceedings{garimella2026dyno,
  title={DYNO: Dynamic Neurosymbolic Orchestrator for Multi-Agent Systems},
  author={Garimella, Ritvik and Shyalika, Chathurangi and Mana, Renjith Prasad Kaippilly and Sheth, Amit},
  booktitle={LLM-based Multi-Agent Systems: Towards Responsible, Reliable, and Scalable Agentic Systems},
  year={2026}
}

@article{sheth2025composite,
  title={Composite AI With Custom, Compact, Neurosymbolic Models: The Emergent Enterprise Artificial Intelligence Paradigm},
  author={Sheth, Amit P and Roy, Kaushik and Venkataramanan, Revathy and Nadimuthu, Venkatesan and Shyalika, Chathurangi},
  journal={IEEE Internet Computing},
  volume={29},
  number={2},
  pages={37--49},
  year={2025},
  publisher={IEEE}
}

@software{rapidfuzz,
  author       = {Max Bachmann},
  title        = {rapidfuzz/RapidFuzz: Release 3.13.0},
  month        = apr,
  year         = 2025,
  publisher    = {Zenodo},
  version      = {v3.13.0},
  doi          = {10.5281/zenodo.15133267},
  url          = {https://doi.org/10.5281/zenodo.15133267},
}

@article{grattafiori2024llama,
  title={The llama 3 herd of models},
  author={Grattafiori, Aaron and Dubey, Abhimanyu and Jauhri, Abhinav and Pandey, Abhinav and Kadian, Abhishek and Al-Dahle, Ahmad and Letman, Aiesha and Mathur, Akhil and Schelten, Alan and Vaughan, Alex and others},
  journal={arXiv preprint arXiv:2407.21783},
  year={2024}
}

@article{liu2026ministral,
  title={Ministral 3},
  author={Liu, Alexander H and Khandelwal, Kartik and Subramanian, Sandeep and Jouault, Victor and Rastogi, Abhinav and Sad{\'e}, Adrien and Jeffares, Alan and Jiang, Albert and Cahill, Alexandre and Gavaudan, Alexandre and others},
  journal={arXiv preprint arXiv:2601.08584},
  year={2026}
}

@article{mishra2024granite,
  title={Granite code models: A family of open foundation models for code intelligence},
  author={Mishra, Mayank and Stallone, Matt and Zhang, Gaoyuan and Shen, Yikang and Prasad, Aditya and Soria, Adriana Meza and Merler, Michele and Selvam, Parameswaran and Surendran, Saptha and Singh, Shivdeep and others},
  journal={arXiv preprint arXiv:2405.04324},
  year={2024}
}

@inproceedings{li2023apibank,
  title     = {{API}-Bank: A Comprehensive Benchmark for Tool-Augmented {LLM}s},
  author    = {Li, Minghao and Zhao, Yingxiu and Yu, Bowen and Song, Feifan and Li, Hangyu and Yu, Haiyang and Li, Zhoujun and Huang, Fei and Li, Yongbin},
  booktitle = {Proceedings of the 2023 Conference on Empirical Methods in Natural Language Processing},
  pages     = {3102--3116},
  year      = {2023},
  publisher = {Association for Computational Linguistics},
  address   = {Singapore},
  doi       = {10.18653/v1/2023.emnlp-main.187},
  url       = {https://aclanthology.org/2023.emnlp-main.187/}
}

@inproceedings{qin2024toolllm,
  title     = {ToolLLM: Facilitating Large Language Models to Master 16000+ Real-world APIs},
  author    = {Qin, Yujia and Liang, Shihao and Ye, Yining and Zhu, Kunlun and Yan, Lan and Lu, Yaxi and Lin, Yankai and Cong, Xin and Tang, Xiangru and Qian, Bill and Zhao, Sihan and Hong, Lauren and Tian, Runchu and Xie, Ruobing and Zhou, Jie and Gerstein, Mark and Li, Dahai and Liu, Zhiyuan and Sun, Maosong},
  booktitle = {The Twelfth International Conference on Learning Representations},
  year      = {2024},
  url       = {https://openreview.net/forum?id=dHng2O0Jjr}
}

@inproceedings{patil2024gorilla,
  title     = {Gorilla: Large Language Model Connected with Massive APIs},
  author    = {Patil, Shishir G. and Zhang, Tianjun and Wang, Xin and Gonzalez, Joseph E.},
  booktitle = {Advances in Neural Information Processing Systems},
  volume    = {37},
  year      = {2024},
  url       = {https://proceedings.neurips.cc/paper_files/paper/2024/hash/e4c61f578ff07830f5c37378dd3ecb0d-Abstract-Conference.html}
}

@inproceedings{patil2025bfcl,
  title     = {The Berkeley Function Calling Leaderboard ({BFCL}): From Tool Use to Agentic Evaluation of Large Language Models},
  author    = {Patil, Shishir G and Mao, Huanzhi and Yan, Fanjia and Ji, Charlie Cheng-Jie and Suresh, Vishnu and Stoica, Ion and Gonzalez, Joseph E.},
  booktitle = {Proceedings of the 42nd International Conference on Machine Learning},
  series    = {Proceedings of Machine Learning Research},
  volume    = {267},
  pages     = {48371--48392},
  year      = {2025},
  publisher = {PMLR},
  url       = {https://proceedings.mlr.press/v267/patil25a.html}
}

@inproceedings{xu2025reducing,
  title     = {Reducing Tool Hallucination via Reliability Alignment},
  author    = {Xu, Hongshen and Zhu, Zichen and Pan, Lei and Wang, Zihan and Zhu, Su and Ma, Da and Cao, Ruisheng and Chen, Lu and Yu, Kai},
  booktitle = {Proceedings of the 42nd International Conference on Machine Learning},
  series    = {Proceedings of Machine Learning Research},
  volume    = {267},
  pages     = {69992--70006},
  year      = {2025},
  publisher = {PMLR},
  url       = {https://proceedings.mlr.press/v267/xu25ap.html}
}

@misc{zhang2024toolbehonest,
  title         = {ToolBeHonest: A Multi-level Hallucination Diagnostic Benchmark for Tool-Augmented Large Language Models},
  author        = {Zhang, Yuxiang and Chen, Jing and Wang, Junjie and Liu, Yaxin and Yang, Cheng and Shi, Chufan and Zhu, Xinyu and Lin, Zihao and Wan, Hanwen and Yang, Yujiu and Sakai, Tetsuya and Feng, Tian and Yamana, Hayato},
  year          = {2024},
  eprint        = {2406.20015},
  archivePrefix = {arXiv},
  primaryClass  = {cs.CL},
  doi           = {10.48550/arXiv.2406.20015},
  url           = {https://arxiv.org/abs/2406.20015}
}

@inproceedings{yao2022react,
  title={React: Synergizing reasoning and acting in language models},
  author={Yao, Shunyu and Zhao, Jeffrey and Yu, Dian and Du, Nan and Shafran, Izhak and Narasimhan, Karthik R and Cao, Yuan},
  booktitle={The eleventh international conference on learning representations}
}

@misc{glaiveai2023functioncalling,
  author       = {{Glaive AI}},
  title        = {glaive-function-calling-v2},
  year         = {2023},
  publisher    = {Hugging Face},
  howpublished = {\url{https://huggingface.co/datasets/glaiveai/glaive-function-calling-v2}},
}

@misc{qian2025toolrlrewardtoollearning,
      title={ToolRL: Reward is All Tool Learning Needs}, 
      author={Cheng Qian and Emre Can Acikgoz and Qi He and Hongru Wang and Xiusi Chen and Dilek Hakkani-Tür and Gokhan Tur and Heng Ji},
      year={2025},
      eprint={2504.13958},
      archivePrefix={arXiv},
      primaryClass={cs.LG},
      url={https://arxiv.org/abs/2504.13958}, 
}

@misc{masterman2024landscapeemergingaiagent,
      title={The Landscape of Emerging AI Agent Architectures for Reasoning, Planning, and Tool Calling: A Survey}, 
      author={Tula Masterman and Sandi Besen and Mason Sawtell and Alex Chao},
      year={2024},
      eprint={2404.11584},
      archivePrefix={arXiv},
      primaryClass={cs.AI},
      url={https://arxiv.org/abs/2404.11584}, 
}

\appendix

\newpage
\appendix

\section{Reproducibility Statement}
\label{app:reproducibility}
All experiments use the following exact model identifiers via Ollama: 
\texttt{llama3.2:3b}, \texttt{ministral:3b}, and \texttt{granite4:3b}. 
Inference is performed at temperature $\tau = 0$ with no system-level 
sampling modifications. The correction loop uses a maximum of 
$T_\text{max} = 15$ iterations. All fuzzy matching thresholds are fixed: 
the parameter drift threshold is $\gamma_\text{pds} = 0.8$; the hard-drift 
boundary for Stage~3 correction is PartialRatio $< 0.6$. All fix templates 
($\psi_1$, $\psi_2$, $\psi_3$) are deterministic and templated: given the 
same failure mode and metric values, the same correction string is produced. 
The Glaive function-calling dataset was filtered to retain only single-call 
rows with well-formed JSON schemas; distractors are sampled uniformly at 
random from the function pool. Experiments were run on a single NVIDIA A100 
GPU. Detailed prompt templates and the full dataset construction pipeline 
will be provided in the supplementary code repository upon acceptance.

\section{Ethics Statement}
\label{app:ethics}
This work proposes a diagnostic framework for evaluating and repairing tool 
calls at inference time. We note several ethical considerations. First, while 
structured feedback reduces argument hallucination, it can also induce 
overcorrection---particularly in function name selection---and should not be 
deployed without monitoring for this failure mode. Second, our evaluation is 
restricted to single-tool selection from a fixed registry using synthetic 
queries; we do not evaluate real tool execution, multi-agent deployment, or 
user-facing applications. Claims about deployment readiness should not be 
inferred from these results. Third, the lexical grounding metric (QSLO) 
captures surface overlap, not semantic correctness; values that are 
semantically valid but lexically distant from the query may be incorrectly 
penalised. Users of this framework should be aware of this limitation when 
interpreting grounding scores.



\section{Dataset Construction}
\label{app:dataset}

We construct a controlled evaluation benchmark from the Glaive
function-calling dataset~\citep{glaiveai2023functioncalling}.
The original corpus is restructured along three axes:
(i)~we retain only rows that require a tool call;
(ii)~we standardise the system prompt to expose each function as a
standalone JSON schema block with typed parameters and explicit
\texttt{required} fields; and
(iii)~we stratify the registry scale to $S \in \{5, 10, 15\}$
functions per row by retaining the ground-truth relevant function and
sampling $S{-}1$ distractors uniformly from the broader function pool.
The same 3{,}000 queries are used across all three registry scales,
so any performance difference across $S$ reflects registry complexity
alone rather than query distribution shift.

Unlike standard function-calling benchmarks that expose only the
relevant tool, our construction forces the model to perform genuine
registry selection: the correct function must be identified from among
$S{-}1$ structurally similar distractors whose parameter types and
description phrasing overlap with the query. As $S$ grows, the risk
of name confusability and argument drift increases. This directly probes
the failure modes captured by our SAAG metrics.

Figure~\ref{fig:sysprompt} shows a representative system prompt at
$S{=}5$. At $S{=}10$ and $S{=}15$, five and ten further distractor
blocks are appended; the query and ground truth are unchanged. The only place ground truth is used is during the extrinsic metric calculation.

\begin{figure}[h]
\begin{tcolorbox}[title=System prompt with agent scale depth as 5]
\ttfamily\scriptsize
SYSTEM: You are a helpful assistant with access to the following
functions. Use them if required -

\{"name": "get\_definition", "description": "Get the definition of a word",
 "parameters": \{"type": "object", "properties": \{"word": \{"type": "string",
 "description": "The word to get the definition of"\}\},
 "required": ["word"]\}\}

\{"name": "generate\_qr\_code", "description": "Generate a QR code for a given text",
 "parameters": \{"type": "object", "properties": \{"text": \{"type": "string",
 "description": "The text to encode"\}, "size": \{"type": "integer",
 "description": "QR code size in pixels"\}\}, "required": ["text"]\}\}

\{"name": "convert\_temperature", "description": "Convert temperature between units",
 "parameters": \{"type": "object", "properties": \{"temperature": \{"type": "number"\},
 "from\_unit": \{"type": "string"\}, "to\_unit": \{"type": "string"\}\},
 "required": ["temperature", "from\_unit", "to\_unit"]\}\}

\{"name": "calculate\_mortgage", "description": "Calculate monthly mortgage payments",
 "parameters": \{"type": "object", "properties": \{"loan\_amount": \{"type": "number"\},
 "interest\_rate": \{"type": "number"\}, "loan\_term": \{"type": "integer"\}\},
 "required": ["loan\_amount", "interest\_rate", "loan\_term"]\}\}

\{"name": "calculate\_tip", "description": "Calculate the tip amount based on the bill",
 "parameters": \{"type": "object", "properties": \{"bill\_amount": \{"type": "number"\},
 "tip\_percentage": \{"type": "number"\}\},
 "required": ["bill\_amount", "tip\_percentage"]\}\}

USER: What does the word `serendipity' mean?

GROUND TRUTH: \{"name": "get\_definition", "arguments": \{"word": "serendipity"\}\}
\end{tcolorbox}
\caption{System prompt at $S{=}5$ for a representative query. The registry
  contains one relevant function (\texttt{get\_definition}) and four distractors.
  At $S{=}10$ and $S{=}15$, five and ten further distractor blocks are appended;
  the query and ground truth remain identical.}
\label{fig:sysprompt}
\end{figure}

\newpage
\section{Structured Correction Messages}
\label{app:fixes}
Table~\ref{tab:fixes} shows representative fix messages $\psi_k$ for each 
failure mode across the three stages. All fixes are constructed 
programmatically from the evaluation output; no human authoring is involved. 
Crucially, no fix reveals the ground-truth function name, parameter set, or 
argument values, ensuring that corrections are grounded solely in the registry 
schema and the model's own prediction.

\begin{table*}[!htbp]
\centering
\small
\renewcommand{\arraystretch}{1.4}
\begin{tabular}{p{1.8cm} p{2.8cm} p{7.8cm}}
\toprule
\textbf{Stage} & \textbf{Failure Mode} & \textbf{Representative Fix Message (Fix Prompts)} \\
\midrule
Q1 & No name found      & \textit{No function name found. Please return only the agent invocation in the exact JSON format specified in the registry.} \\
Q1 & Wrong name         & \textit{Function name ``\{predicted\}'' not found in registry. Please check the system prompt for the correct function names and respond with only the agent invocation in the exact JSON format specified.} \\
\midrule
Q2 & Missing required   & \textit{Key parameters of the agent from the registry are missing. Please ensure the invocation includes all parameters specified for that agent in the system prompt.} \\
Q2 & Spurious params    & \textit{Spurious parameters not present in the registry are included. Please include only the parameters specified for that agent.} \\
Q2 & Drifted params     & \textit{Some parameters appear misspelled and deviate slightly from the registry. Please ensure all parameter names exactly match the schema in the system prompt.} \\
\midrule
Q3 & Numeric mismatch   & \textit{The numeric values for \{params\} do not match what the user stated. Use the exact numbers from the user's request.} \\
Q3 & Hard value drift   & \textit{The values for \{params\} have no grounding in the user's request. Do not invent values. Use only what the user explicitly stated, or omit the argument.} \\
Q3 & Soft value drift   & \textit{The values for \{params\} are loosely related but not exact. Use the precise wording or value from the user's request.} \\
\bottomrule
\end{tabular}
\caption{Representative structured correction messages $\psi_k$ per failure 
mode. Bracketed tokens (\{predicted\}, \{params\}) are instantiated 
programmatically from the evaluation output.}
\label{tab:fixes}
\end{table*}

\section{Per-Scale Results}
\label{app:res}

Per-scale breakdowns of intrinsic grounding metrics and extrinsic 
ground-truth metrics for all three models across registry scales 
$S \in \{5, 10, 15\}$; summary tables averaged across scales are reported 
in the main paper (intrinsic: Table~\ref{tab:intrinsic}; extrinsic: 
Table~\ref{tab:extrinsic}). Tables~\ref{tab:llama_s5}--\ref{tab:granite_agg} 
present full per-scale breakdowns, with each table reporting intrinsic 
grounding metrics (top block) and extrinsic correctness metrics (bottom 
block) as mean\,$\pm$\,std over all rows at that registry scale and the respective solve rate.


\begin{table*}[t]
\centering
\setlength{\tabcolsep}{5pt}
\renewcommand{\arraystretch}{1.2}
\footnotesize
\begin{tabular}{@{}ll ccc@{}}
\toprule
\textbf{Group} & \textbf{Metric} & \textbf{Direct} & \textbf{Binary} & \textbf{SAAG} \\
\midrule
\multirow{8}{*}{\textsc{Intrinsic}} & FNEM $\uparrow$ & 1.000{\tiny$\pm$0.000} & 0.999{\tiny$\pm$0.026} & 1.000{\tiny$\pm$0.000} \\
 & NCS $\uparrow$ & 1.000{\tiny$\pm$0.000} & 0.999{\tiny$\pm$0.026} & 1.000{\tiny$\pm$0.000} \\
 & RPR $\uparrow$ & 0.979{\tiny$\pm$0.136} & 0.970{\tiny$\pm$0.147} & 0.976{\tiny$\pm$0.131} \\
 & PDS $\uparrow$ & 0.006{\tiny$\pm$0.075} & 0.002{\tiny$\pm$0.037} & 0.001{\tiny$\pm$0.033} \\
 & SPR $\downarrow$ & 0.048{\tiny$\pm$0.208} & 0.008{\tiny$\pm$0.080} & 0.002{\tiny$\pm$0.039} \\
 & AVEM $\uparrow$ & 0.567{\tiny$\pm$0.438} & 0.620{\tiny$\pm$0.446} & 0.653{\tiny$\pm$0.441} \\
 & QSLO $\uparrow$ & 0.325{\tiny$\pm$0.233} & 0.340{\tiny$\pm$0.239} & 0.334{\tiny$\pm$0.242} \\
 & VHR $\downarrow$ & 0.351{\tiny$\pm$0.381} & 0.312{\tiny$\pm$0.401} & 0.288{\tiny$\pm$0.396} \\
\midrule
\multirow{3}{*}{\textsc{Extrinsic}} & 
Precision $\uparrow$ & 0.561{\tiny$\pm$0.451} & 0.556{\tiny$\pm$0.466} & 0.569{\tiny$\pm$0.465} \\
 & Recall $\uparrow$ & 0.573{\tiny$\pm$0.455} & 0.552{\tiny$\pm$0.465} & 0.561{\tiny$\pm$0.464} \\
 & F1 $\uparrow$ & 0.564{\tiny$\pm$0.450} & 0.550{\tiny$\pm$0.462} & 0.561{\tiny$\pm$0.461} \\
\midrule
\multirow{1}{*}{\textsc{Stats}} & Solve Rate $\uparrow$ & 0.6328 & 0.7835 & 0.8172 \\
\bottomrule
\end{tabular}
\caption{LLaMA 3.2 at registry scale $S=5$. The top panel reports intrinsic grounding metrics, while the bottom panel shows extrinsic correctness metrics across feedback conditions. Values are reported as $\text{mean} \pm \text{std}$ over all rows at this scale; the standard deviation reflects the dispersion of row-level scores within the $[0,1]$ range. Arrows indicate directionality: higher is better ($\uparrow$), lower is better ($\downarrow$).}
\label{tab:llama_s5}
\end{table*}

\begin{table*}[t]
\centering
\setlength{\tabcolsep}{5pt}
\renewcommand{\arraystretch}{1.2}
\footnotesize
\begin{tabular}{@{}ll ccc@{}}
\toprule
\textbf{Group} & \textbf{Metric} & \textbf{Direct} & \textbf{Binary} & \textbf{SAAG} \\
\midrule
\multirow{8}{*}{\textsc{Intrinsic}} & FNEM $\uparrow$ & 0.999{\tiny$\pm$0.037} & 0.997{\tiny$\pm$0.082} & 1.000{\tiny$\pm$0.000} \\
 & NCS $\uparrow$ & 1.000{\tiny$\pm$0.000} & 1.000{\tiny$\pm$0.000} & 1.000{\tiny$\pm$0.000} \\
 & RPR $\uparrow$ & 0.955{\tiny$\pm$0.198} & 0.963{\tiny$\pm$0.164} & 0.970{\tiny$\pm$0.143} \\
 & PDS $\uparrow$ & 0.010{\tiny$\pm$0.097} & 0.005{\tiny$\pm$0.066} & 0.005{\tiny$\pm$0.066} \\
 & SPR $\downarrow$ & 0.068{\tiny$\pm$0.240} & 0.017{\tiny$\pm$0.118} & 0.009{\tiny$\pm$0.090} \\
 & AVEM $\uparrow$ & 0.560{\tiny$\pm$0.439} & 0.618{\tiny$\pm$0.443} & 0.646{\tiny$\pm$0.437} \\
 & QSLO $\uparrow$ & 0.331{\tiny$\pm$0.225} & 0.351{\tiny$\pm$0.238} & 0.358{\tiny$\pm$0.241} \\
 & VHR $\downarrow$ & 0.354{\tiny$\pm$0.380} & 0.310{\tiny$\pm$0.393} & 0.287{\tiny$\pm$0.388} \\
\midrule
\multirow{3}{*}{\textsc{Extrinsic}} & Precision $\uparrow$ & 0.524{\tiny$\pm$0.452} & 0.533{\tiny$\pm$0.465} & 0.547{\tiny$\pm$0.466} \\
 & Recall $\uparrow$ & 0.538{\tiny$\pm$0.457} & 0.532{\tiny$\pm$0.465} & 0.541{\tiny$\pm$0.464} \\
 & F1 $\uparrow$ & 0.528{\tiny$\pm$0.452} & 0.528{\tiny$\pm$0.461} & 0.540{\tiny$\pm$0.462} \\
 \midrule
\multirow{1}{*}{\textsc{Stats}} & Solve Rate $\uparrow$ & 0.6093  & 0.7641 & 0.7995 \\
\bottomrule
\end{tabular}
\caption{LLaMA 3.2 at registry scale $S=10$. The top panel reports intrinsic grounding metrics, while the bottom panel shows extrinsic correctness metrics across feedback conditions. Values are reported as $\text{mean} \pm \text{std}$ over all rows at this scale; the standard deviation reflects the dispersion of row-level scores within the $[0,1]$ range. Arrows indicate directionality: higher is better ($\uparrow$), lower is better ($\downarrow$).}
\label{tab:llama_s10}
\end{table*}

\begin{table*}[t]
\centering
\setlength{\tabcolsep}{5pt}
\renewcommand{\arraystretch}{1.2}
\footnotesize
\begin{tabular}{@{}ll ccc@{}}
\toprule
\textbf{Group} & \textbf{Metric} & \textbf{Direct} & \textbf{Binary} & \textbf{SAAG} \\
\midrule
\multirow{8}{*}{\textsc{Intrinsic}} & FNEM $\uparrow$ & 1.000{\tiny$\pm$0.000} & 0.997{\tiny$\pm$0.082} & 1.000{\tiny$\pm$0.000} \\
 & NCS $\uparrow$ & 1.000{\tiny$\pm$0.000} & 1.000{\tiny$\pm$0.000} & 1.000{\tiny$\pm$0.000} \\
 & RPR $\uparrow$ & 0.969{\tiny$\pm$0.149} & 0.962{\tiny$\pm$0.165} & 0.944{\tiny$\pm$0.217} \\
 & PDS $\uparrow$ & 0.008{\tiny$\pm$0.087} & 0.009{\tiny$\pm$0.090} & 0.019{\tiny$\pm$0.131} \\
 & SPR $\downarrow$ & 0.013{\tiny$\pm$0.103} & 0.019{\tiny$\pm$0.123} & 0.073{\tiny$\pm$0.246} \\
 & AVEM $\uparrow$ & 0.625{\tiny$\pm$0.440} & 0.601{\tiny$\pm$0.444} & 0.552{\tiny$\pm$0.437} \\
 & QSLO $\uparrow$ & 0.361{\tiny$\pm$0.240} & 0.350{\tiny$\pm$0.231} & 0.336{\tiny$\pm$0.232} \\
 & VHR $\downarrow$ & 0.303{\tiny$\pm$0.392} & 0.321{\tiny$\pm$0.393} & 0.359{\tiny$\pm$0.379} \\
\midrule
\multirow{3}{*}{\textsc{Extrinsic}} & Precision $\uparrow$ & 0.488{\tiny$\pm$0.451} & 0.487{\tiny$\pm$0.464} & 0.503{\tiny$\pm$0.465} \\
 & Recall $\uparrow$ & 0.499{\tiny$\pm$0.456} & 0.488{\tiny$\pm$0.465} & 0.500{\tiny$\pm$0.465} \\
 & F1 $\uparrow$ & 0.490{\tiny$\pm$0.450} & 0.485{\tiny$\pm$0.461} & 0.498{\tiny$\pm$0.461} \\
 \midrule
\multirow{1}{*}{\textsc{Stats}} & Solve Rate $\uparrow$ & 0.6015 & 0.7559 & 0.7889 \\
\bottomrule
\end{tabular}
\caption{LLaMA 3.2 at registry scale $S=15$. The top panel reports intrinsic grounding metrics, while the bottom panel shows extrinsic correctness metrics across feedback conditions. Values are reported as $\text{mean} \pm \text{std}$ over all rows at this scale; the standard deviation reflects the dispersion of row-level scores within the $[0,1]$ range. Arrows indicate directionality: higher is better ($\uparrow$), lower is better ($\downarrow$).}
\label{tab:llama_s15}
\end{table*}

\begin{table*}[t]
\centering
\setlength{\tabcolsep}{5pt}
\renewcommand{\arraystretch}{1.2}
\footnotesize
\begin{tabular}{@{}ll ccc@{}}
\toprule
\textbf{Group} & \textbf{Metric} & \textbf{Direct} & \textbf{Binary} & \textbf{SAAG} \\
\midrule
\multirow{8}{*}{\textsc{Intrinsic}} & FNEM $\uparrow$ & 0.9997 $\pm$ 0.0006 & 0.9977 $\pm$ 0.0012 & 1.0000 $\pm$ 0.0000 \\
 & NCS $\uparrow$ & 1.0000 $\pm$ 0.0000 & 0.9997 $\pm$ 0.0006 & 1.0000 $\pm$ 0.0000 \\
 & RPR $\uparrow$ & 0.9677 $\pm$ 0.0121 & 0.9650 $\pm$ 0.0044 & 0.9633 $\pm$ 0.0173 \\
 & PDS $\uparrow$ & 0.0080 $\pm$ 0.0020 & 0.0053 $\pm$ 0.0035 & 0.0083 $\pm$ 0.0095 \\
 & SPR $\downarrow$ & 0.0430 $\pm$ 0.0283 & 0.0147 $\pm$ 0.0059 & 0.0280 $\pm$ 0.0392 \\
 & AVEM $\uparrow$ & 0.5840 $\pm$ 0.0359 & 0.6130 $\pm$ 0.0104 & 0.6170 $\pm$ 0.0525 \\
 & QSLO $\uparrow$ & 0.3390 $\pm$ 0.0187 & 0.3470 $\pm$ 0.0061 & 0.3427 $\pm$ 0.0122 \\
 & VHR $\downarrow$ & 0.3360 $\pm$ 0.0284 & 0.3143 $\pm$ 0.0059 & 0.3113 $\pm$ 0.0406 \\
\midrule
\multirow{3}{*}{\textsc{Extrinsic}} & Precision $\uparrow$ & 0.5243 $\pm$ 0.0366 & 0.5253 $\pm$ 0.0349 & 0.5397 $\pm$ 0.0330 \\
 & Recall $\uparrow$ & 0.5367 $\pm$ 0.0370 & 0.5240 $\pm$ 0.0333 & 0.5340 $\pm$ 0.0313 \\
 & F1 $\uparrow$ & 0.5273 $\pm$ 0.0370 & 0.5210 $\pm$ 0.0326 & 0.5330 $\pm$ 0.0322 \\
\midrule
\multirow{1}{*}{\textsc{Stats}} & Solve Rate $\uparrow$ & 0.6145 $\pm$ 0.0160 & 0.7678 $\pm$ 0.0140 & 0.8019 $\pm$ 0.0145 \\
\bottomrule
\end{tabular}
\caption{LLaMA 3.2 aggregated across registry scales $S=5,10,15$. Values are reported as mean $\pm$ sample standard deviation over the three scale-level results.}
\label{tab:llama_agg}
\end{table*}


\begin{table*}[t]
\centering
\setlength{\tabcolsep}{5pt}
\renewcommand{\arraystretch}{1.2}
\footnotesize
\begin{tabular}{@{}ll ccc@{}}
\toprule
\textbf{Group} & \textbf{Metric} & \textbf{Direct} & \textbf{Binary} & \textbf{SAAG} \\
\midrule
\multirow{8}{*}{\textsc{Intrinsic}} & FNEM $\uparrow$ & 1.000{\tiny$\pm$0.000} & 1.000{\tiny$\pm$0.000} & 0.999{\tiny$\pm$0.041} \\
 & NCS $\uparrow$ & 1.000{\tiny$\pm$0.000} & 1.000{\tiny$\pm$0.000} & 1.000{\tiny$\pm$0.011} \\
 & RPR $\uparrow$ & 0.987{\tiny$\pm$0.112} & 0.934{\tiny$\pm$0.248} & 0.877{\tiny$\pm$0.321} \\
 & PDS $\uparrow$ & 0.000{\tiny$\pm$0.016} & 0.000{\tiny$\pm$0.016} & 0.000{\tiny$\pm$0.016} \\
 & SPR $\downarrow$ & 0.003{\tiny$\pm$0.049} & 0.000{\tiny$\pm$0.013} & 0.001{\tiny$\pm$0.032} \\
 & AVEM $\uparrow$ & 0.662{\tiny$\pm$0.433} & 0.717{\tiny$\pm$0.425} & 0.791{\tiny$\pm$0.377} \\
 & QSLO $\uparrow$ & 0.279{\tiny$\pm$0.264} & 0.343{\tiny$\pm$0.240} & 0.369{\tiny$\pm$0.323} \\
 & VHR $\downarrow$ & 0.271{\tiny$\pm$0.388} & 0.215{\tiny$\pm$0.375} & 0.164{\tiny$\pm$0.345} \\
\midrule
\multirow{3}{*}{\textsc{Extrinsic}} & Precision $\uparrow$ & 0.718{\tiny$\pm$0.427} & 0.660{\tiny$\pm$0.456} & 0.674{\tiny$\pm$0.448} \\
 & Recall $\uparrow$ & 0.723{\tiny$\pm$0.428} & 0.660{\tiny$\pm$0.456} & 0.668{\tiny$\pm$0.447} \\
 & F1 $\uparrow$ & 0.718{\tiny$\pm$0.426} & 0.658{\tiny$\pm$0.454} & 0.668{\tiny$\pm$0.446} \\
\midrule
\multirow{1}{*}{\textsc{Stats}} & Solve Rate $\uparrow$ & 0.7318{\tiny$\pm$0.000} & 0.8286{\tiny$\pm$0.000} & 0.8199{\tiny$\pm$0.000} \\
\bottomrule
\end{tabular}
\caption{Ministral 3 at registry scale $S=5$. The top panel reports intrinsic grounding metrics, while the bottom panel shows extrinsic correctness metrics across feedback conditions. Values are reported as $\text{mean} \pm \text{std}$ over all rows at this scale; the standard deviation reflects the dispersion of row-level scores within the $[0,1]$ range. Arrows indicate directionality: higher is better ($\uparrow$), lower is better ($\downarrow$).}
\label{tab:ministral_s5}
\end{table*}

\begin{table*}[t]
\centering
\setlength{\tabcolsep}{5pt}
\renewcommand{\arraystretch}{1.2}
\footnotesize
\begin{tabular}{@{}ll ccc@{}}
\toprule
\textbf{Group} & \textbf{Metric} & \textbf{Direct} & \textbf{Binary} & \textbf{SAAG} \\
\midrule
\multirow{8}{*}{\textsc{Intrinsic}} & FNEM $\uparrow$ & 1.000{\tiny$\pm$0.000} & 1.000{\tiny$\pm$0.000} & 1.000{\tiny$\pm$0.000} \\
 & NCS $\uparrow$ & 1.000{\tiny$\pm$0.000} & 1.000{\tiny$\pm$0.000} & 1.000{\tiny$\pm$0.000} \\
 & RPR $\uparrow$ & 0.986{\tiny$\pm$0.113} & 0.938{\tiny$\pm$0.240} & 0.887{\tiny$\pm$0.309} \\
 & PDS $\uparrow$ & 0.001{\tiny$\pm$0.023} & 0.002{\tiny$\pm$0.037} & 0.001{\tiny$\pm$0.033} \\
 & SPR $\downarrow$ & 0.006{\tiny$\pm$0.074} & 0.002{\tiny$\pm$0.040} & 0.002{\tiny$\pm$0.039} \\
 & AVEM $\uparrow$ & 0.658{\tiny$\pm$0.435} & 0.720{\tiny$\pm$0.421} & 0.792{\tiny$\pm$0.376} \\
 & QSLO $\uparrow$ & 0.266{\tiny$\pm$0.275} & 0.326{\tiny$\pm$0.253} & 0.392{\tiny$\pm$0.330} \\
 & VHR $\downarrow$ & 0.277{\tiny$\pm$0.394} & 0.209{\tiny$\pm$0.369} & 0.159{\tiny$\pm$0.338} \\
\midrule
\multirow{3}{*}{\textsc{Extrinsic}}  & Precision $\uparrow$ & 0.705{\tiny$\pm$0.434} & 0.650{\tiny$\pm$0.461} & 0.659{\tiny$\pm$0.455} \\
 & Recall $\uparrow$ & 0.709{\tiny$\pm$0.435} & 0.649{\tiny$\pm$0.460} & 0.653{\tiny$\pm$0.454} \\
 & F1 $\uparrow$ & 0.705{\tiny$\pm$0.433} & 0.648{\tiny$\pm$0.459} & 0.653{\tiny$\pm$0.452} \\
\midrule
\multirow{1}{*}{\textsc{Stats}} & Solve Rate $\uparrow$ & 0.7291{\tiny$\pm$0.000} & 0.8285{\tiny$\pm$0.000} & 0.8292{\tiny$\pm$0.000} \\
\bottomrule
\end{tabular}
\caption{Ministral 3 at registry scale $S=10$. The top panel reports intrinsic grounding metrics, while the bottom panel shows extrinsic correctness metrics across feedback conditions. Values are reported as $\text{mean} \pm \text{std}$ over all rows at this scale; the standard deviation reflects the dispersion of row-level scores within the $[0,1]$ range. Arrows indicate directionality: higher is better ($\uparrow$), lower is better ($\downarrow$).}
\label{tab:ministral_s10}
\end{table*}

\begin{table*}[t]
\centering
\setlength{\tabcolsep}{5pt}
\renewcommand{\arraystretch}{1.2}
\footnotesize
\begin{tabular}{@{}ll ccc@{}}
\toprule
\textbf{Group} & \textbf{Metric} & \textbf{Direct} & \textbf{Binary} & \textbf{SAAG} \\
\midrule
\multirow{8}{*}{\textsc{Intrinsic}} & FNEM $\uparrow$ & 1.000{\tiny$\pm$0.000} & 1.000{\tiny$\pm$0.000} & 1.000{\tiny$\pm$0.000} \\
 & NCS $\uparrow$ & 1.000{\tiny$\pm$0.000} & 1.000{\tiny$\pm$0.000} & 1.000{\tiny$\pm$0.000} \\
 & RPR $\uparrow$ & 0.978{\tiny$\pm$0.142} & 0.938{\tiny$\pm$0.238} & 0.893{\tiny$\pm$0.302} \\
 & PDS $\uparrow$ & 0.003{\tiny$\pm$0.055} & 0.002{\tiny$\pm$0.047} & 0.002{\tiny$\pm$0.038} \\
 & SPR $\downarrow$ & 0.014{\tiny$\pm$0.112} & 0.002{\tiny$\pm$0.037} & 0.003{\tiny$\pm$0.057} \\
 & AVEM $\uparrow$ & 0.654{\tiny$\pm$0.437} & 0.714{\tiny$\pm$0.425} & 0.776{\tiny$\pm$0.389} \\
 & QSLO $\uparrow$ & 0.259{\tiny$\pm$0.264} & 0.323{\tiny$\pm$0.249} & 0.323{\tiny$\pm$0.336} \\
 & VHR $\downarrow$ & 0.283{\tiny$\pm$0.397} & 0.216{\tiny$\pm$0.374} & 0.174{\tiny$\pm$0.352} \\
\midrule
\multirow{3}{*}{\textsc{Extrinsic}} & Precision $\uparrow$ & 0.693{\tiny$\pm$0.437} & 0.635{\tiny$\pm$0.463} & 0.649{\tiny$\pm$0.458} \\
 & Recall $\uparrow$ & 0.697{\tiny$\pm$0.438} & 0.634{\tiny$\pm$0.464} & 0.642{\tiny$\pm$0.457} \\
 & F1 $\uparrow$ & 0.692{\tiny$\pm$0.436} & 0.632{\tiny$\pm$0.461} & 0.642{\tiny$\pm$0.455} \\
\midrule
\multirow{1}{*}{\textsc{Stats}} & Solve Rate $\uparrow$ & 0.7169{\tiny$\pm$0.000} & 0.8293{\tiny$\pm$0.000} & 0.8263{\tiny$\pm$0.000} \\
\bottomrule
\end{tabular}
\caption{Ministral 3 at registry scale $S=15$. The top panel reports intrinsic grounding metrics, while the bottom panel shows extrinsic correctness metrics across feedback conditions. Values are reported as $\text{mean} \pm \text{std}$ over all rows at this scale; the standard deviation reflects the dispersion of row-level scores within the $[0,1]$ range. Arrows indicate directionality: higher is better ($\uparrow$), lower is better ($\downarrow$).}
\label{tab:ministral_s15}
\end{table*}

\begin{table*}[t]
\centering
\setlength{\tabcolsep}{5pt}
\renewcommand{\arraystretch}{1.2}
\footnotesize
\begin{tabular}{@{}ll ccc@{}}
\toprule
\textbf{Group} & \textbf{Metric} & \textbf{Direct} & \textbf{Binary} & \textbf{SAAG} \\
\midrule
\multirow{8}{*}{\textsc{Intrinsic}} & FNEM $\uparrow$ & 1.0000 $\pm$ 0.0000 & 1.0000 $\pm$ 0.0000 & 0.9997 $\pm$ 0.0006 \\
 & NCS $\uparrow$ & 1.0000 $\pm$ 0.0000 & 1.0000 $\pm$ 0.0000 & 1.0000 $\pm$ 0.0000 \\
 & RPR $\uparrow$ & 0.9837 $\pm$ 0.0049 & 0.9367 $\pm$ 0.0023 & 0.8857 $\pm$ 0.0081 \\
 & PDS $\uparrow$ & 0.0013 $\pm$ 0.0015 & 0.0013 $\pm$ 0.0012 & 0.0010 $\pm$ 0.0010 \\
 & SPR $\downarrow$ & 0.0077 $\pm$ 0.0057 & 0.0013 $\pm$ 0.0012 & 0.0020 $\pm$ 0.0010 \\
 & AVEM $\uparrow$ & 0.6580 $\pm$ 0.0040 & 0.7170 $\pm$ 0.0030 & 0.7863 $\pm$ 0.0090 \\
 & QSLO $\uparrow$ & 0.2680 $\pm$ 0.0101 & 0.3307 $\pm$ 0.0108 & 0.3613 $\pm$ 0.0351 \\
 & VHR $\downarrow$ & 0.2770 $\pm$ 0.0060 & 0.2133 $\pm$ 0.0038 & 0.1657 $\pm$ 0.0076 \\
\midrule
\multirow{3}{*}{\textsc{Extrinsic}} & Precision $\uparrow$ & 0.7053 $\pm$ 0.0125 & 0.6483 $\pm$ 0.0126 & 0.6607 $\pm$ 0.0126 \\
 & Recall $\uparrow$ & 0.7097 $\pm$ 0.0130 & 0.6477 $\pm$ 0.0131 & 0.6543 $\pm$ 0.0131 \\
 & F1 $\uparrow$ & 0.7050 $\pm$ 0.0130 & 0.6460 $\pm$ 0.0131 & 0.6543 $\pm$ 0.0131 \\
\midrule
\multirow{1}{*}{\textsc{Stats}} & Solve Rate $\uparrow$ & 0.7259 $\pm$ 0.0079 & 0.8288 $\pm$ 0.0004 & 0.8251 $\pm$ 0.0048 \\
\bottomrule
\end{tabular}
\caption{Ministral 3 aggregated across registry scales $S=5,10,15$. Values are reported as mean $\pm$ sample standard deviation over the three scale-level results.}
\label{tab:ministral_agg}
\end{table*}


\begin{table*}[t]
\centering
\setlength{\tabcolsep}{5pt}
\renewcommand{\arraystretch}{1.2}
\footnotesize
\begin{tabular}{@{}ll ccc@{}}
\toprule
\textbf{Group} & \textbf{Metric} & \textbf{Direct} & \textbf{Binary} & \textbf{SAAG} \\
\midrule
\multirow{8}{*}{\textsc{Intrinsic}} & FNEM $\uparrow$ & 1.000{\tiny$\pm$0.000} & 1.000{\tiny$\pm$0.000} & 1.000{\tiny$\pm$0.000} \\
 & NCS $\uparrow$ & 1.000{\tiny$\pm$0.000} & 1.000{\tiny$\pm$0.000} & 1.000{\tiny$\pm$0.000} \\
 & RPR $\uparrow$ & 0.980{\tiny$\pm$0.133} & 0.988{\tiny$\pm$0.100} & 0.918{\tiny$\pm$0.222} \\
 & PDS $\uparrow$ & 0.008{\tiny$\pm$0.084} & 0.004{\tiny$\pm$0.057} & 0.007{\tiny$\pm$0.077} \\
 & SPR $\downarrow$ & 0.016{\tiny$\pm$0.117} & 0.005{\tiny$\pm$0.067} & 0.013{\tiny$\pm$0.107} \\
 & AVEM $\uparrow$ & 0.652{\tiny$\pm$0.437} & 0.646{\tiny$\pm$0.457} & 0.707{\tiny$\pm$0.431} \\
 & QSLO $\uparrow$ & 0.327{\tiny$\pm$0.227} & 0.303{\tiny$\pm$0.235} & 0.286{\tiny$\pm$0.312} \\
 & VHR $\downarrow$ & 0.272{\tiny$\pm$0.377} & 0.288{\tiny$\pm$0.411} & 0.245{\tiny$\pm$0.399} \\
\midrule
\multirow{3}{*}{\textsc{Extrinsic}} & Precision $\uparrow$ & 0.762{\tiny$\pm$0.401} & 0.676{\tiny$\pm$0.454} & 0.750{\tiny$\pm$0.420} \\
 & Recall $\uparrow$ & 0.759{\tiny$\pm$0.402} & 0.672{\tiny$\pm$0.453} & 0.713{\tiny$\pm$0.418} \\
 & F1 $\uparrow$ & 0.758{\tiny$\pm$0.401} & 0.673{\tiny$\pm$0.452} & 0.725{\tiny$\pm$0.415} \\
\midrule
\multirow{1}{*}{\textsc{Stats}} & Solve Rate $\uparrow$ & 0.7151 & 0.8182 & 0.7492 \\
\bottomrule
\end{tabular}
\caption{Granite 4 at registry scale $S=5$. The top panel reports intrinsic grounding metrics, while the bottom panel shows extrinsic correctness metrics across feedback conditions. Values are reported as $\text{mean} \pm \text{std}$ over all rows at this scale; the standard deviation reflects the dispersion of row-level scores within the $[0,1]$ range. Arrows indicate directionality: higher is better ($\uparrow$), lower is better ($\downarrow$).}
\label{tab:granite_d5}
\end{table*}

\begin{table*}[t]
\centering
\setlength{\tabcolsep}{5pt}
\renewcommand{\arraystretch}{1.2}
\footnotesize
\begin{tabular}{@{}ll ccc@{}}
\toprule
\textbf{Group} & \textbf{Metric} & \textbf{Direct} & \textbf{Binary} & \textbf{SAAG} \\
\midrule
\multirow{8}{*}{\textsc{Intrinsic}} & FNEM $\uparrow$ & 1.000{\tiny$\pm$0.000} & 0.999{\tiny$\pm$0.032} & 1.000{\tiny$\pm$0.000} \\
 & NCS $\uparrow$ & 1.000{\tiny$\pm$0.000} & 0.999{\tiny$\pm$0.032} & 1.000{\tiny$\pm$0.000} \\
 & RPR $\uparrow$ & 0.957{\tiny$\pm$0.187} & 0.974{\tiny$\pm$0.151} & 0.896{\tiny$\pm$0.254} \\
 & PDS $\uparrow$ & 0.015{\tiny$\pm$0.117} & 0.008{\tiny$\pm$0.084} & 0.014{\tiny$\pm$0.114} \\
 & SPR $\downarrow$ & 0.030{\tiny$\pm$0.156} & 0.017{\tiny$\pm$0.116} & 0.027{\tiny$\pm$0.149} \\
 & AVEM $\uparrow$ & 0.647{\tiny$\pm$0.437} & 0.648{\tiny$\pm$0.452} & 0.704{\tiny$\pm$0.431} \\
 & QSLO $\uparrow$ & 0.314{\tiny$\pm$0.227} & 0.293{\tiny$\pm$0.231} & 0.294{\tiny$\pm$0.300} \\
 & VHR $\downarrow$ & 0.280{\tiny$\pm$0.380} & 0.286{\tiny$\pm$0.410} & 0.250{\tiny$\pm$0.400} \\
\midrule
\multirow{3}{*}{\textsc{Extrinsic}} & Precision $\uparrow$ & 0.748{\tiny$\pm$0.407} & 0.673{\tiny$\pm$0.455} & 0.740{\tiny$\pm$0.424} \\
 & Recall $\uparrow$ & 0.746{\tiny$\pm$0.408} & 0.669{\tiny$\pm$0.455} & 0.704{\tiny$\pm$0.422} \\
 & F1 $\uparrow$ & 0.745{\tiny$\pm$0.406} & 0.669{\tiny$\pm$0.453} & 0.715{\tiny$\pm$0.418} \\
\midrule
\multirow{1}{*}{\textsc{Stats}} & Solve Rate $\uparrow$ & 0.6910 & 0.8138 & 0.7374 \\
\bottomrule
\end{tabular}
\caption{Granite 4 at registry scale $S=10$. The top panel reports intrinsic grounding metrics, while the bottom panel shows extrinsic correctness metrics across feedback conditions. Values are reported as $\text{mean} \pm \text{std}$ over all rows at this scale; the standard deviation reflects the dispersion of row-level scores within the $[0,1]$ range. Arrows indicate directionality: higher is better ($\uparrow$), lower is better ($\downarrow$).}
\label{tab:granite_d10}
\end{table*}

\begin{table*}[t]
\centering
\setlength{\tabcolsep}{5pt}
\renewcommand{\arraystretch}{1.2}
\footnotesize
\begin{tabular}{@{}ll ccc@{}}
\toprule
\textbf{Group} & \textbf{Metric} & \textbf{Direct} & \textbf{Binary} & \textbf{SAAG} \\
\midrule
\multirow{8}{*}{\textsc{Intrinsic}} & FNEM $\uparrow$ & 1.000{\tiny$\pm$0.000} & 0.996{\tiny$\pm$0.063} & 0.996{\tiny$\pm$0.063} \\
 & NCS $\uparrow$ & 1.000{\tiny$\pm$0.000} & 0.996{\tiny$\pm$0.063} & 0.996{\tiny$\pm$0.063} \\
 & RPR $\uparrow$ & 0.938{\tiny$\pm$0.227} & 0.965{\tiny$\pm$0.175} & 0.964{\tiny$\pm$0.178} \\
 & PDS $\uparrow$ & 0.021{\tiny$\pm$0.135} & 0.012{\tiny$\pm$0.102} & 0.012{\tiny$\pm$0.104} \\
 & SPR $\downarrow$ & 0.046{\tiny$\pm$0.196} & 0.020{\tiny$\pm$0.133} & 0.019{\tiny$\pm$0.128} \\
 & AVEM $\uparrow$ & 0.646{\tiny$\pm$0.438} & 0.642{\tiny$\pm$0.456} & 0.653{\tiny$\pm$0.453} \\
 & QSLO $\uparrow$ & 0.312{\tiny$\pm$0.228} & 0.293{\tiny$\pm$0.229} & 0.294{\tiny$\pm$0.224} \\
 & VHR $\downarrow$ & 0.281{\tiny$\pm$0.381} & 0.289{\tiny$\pm$0.413} & 0.283{\tiny$\pm$0.412} \\
\midrule
\multirow{3}{*}{\textsc{Extrinsic}} & Precision $\uparrow$ & 0.732{\tiny$\pm$0.416} & 0.653{\tiny$\pm$0.461} & 0.661{\tiny$\pm$0.460} \\
 & Recall $\uparrow$ & 0.730{\tiny$\pm$0.417} & 0.648{\tiny$\pm$0.461} & 0.654{\tiny$\pm$0.459} \\
 & F1 $\uparrow$ & 0.729{\tiny$\pm$0.415} & 0.649{\tiny$\pm$0.460} & 0.655{\tiny$\pm$0.458} \\
\midrule
\multirow{1}{*}{\textsc{Stats}} & Solve Rate $\uparrow$ & 0.6749 & 0.8173 & 0.8259 \\
\bottomrule
\end{tabular}
\caption{Granite 4 at registry scale $S=15$. The top panel reports intrinsic grounding metrics, while the bottom panel shows extrinsic correctness metrics across feedback conditions. Values are reported as $\text{mean} \pm \text{std}$ over all rows at this scale; the standard deviation reflects the dispersion of row-level scores within the $[0,1]$ range. Arrows indicate directionality: higher is better ($\uparrow$), lower is better ($\downarrow$).}
\label{tab:granite_d15}
\end{table*}

\begin{table*}[t]
\centering
\setlength{\tabcolsep}{5pt}
\renewcommand{\arraystretch}{1.2}
\footnotesize
\begin{tabular}{@{}ll ccc@{}}
\toprule
\textbf{Group} & \textbf{Metric} & \textbf{Direct} & \textbf{Binary} & \textbf{SAAG} \\
\midrule
\multirow{8}{*}{\textsc{Intrinsic}} & FNEM $\uparrow$ & 1.0000 $\pm$ 0.0000 & 0.9983 $\pm$ 0.0021 & 0.9987 $\pm$ 0.0023 \\
 & NCS $\uparrow$ & 1.0000 $\pm$ 0.0000 & 0.9983 $\pm$ 0.0021 & 0.9987 $\pm$ 0.0023 \\
 & RPR $\uparrow$ & 0.9583 $\pm$ 0.0210 & 0.9757 $\pm$ 0.0116 & 0.9260 $\pm$ 0.0347 \\
 & PDS $\uparrow$ & 0.0147 $\pm$ 0.0065 & 0.0080 $\pm$ 0.0040 & 0.0110 $\pm$ 0.0036 \\
 & SPR $\downarrow$ & 0.0307 $\pm$ 0.0150 & 0.0140 $\pm$ 0.0079 & 0.0197 $\pm$ 0.0070 \\
 & AVEM $\uparrow$ & 0.6483 $\pm$ 0.0032 & 0.6453 $\pm$ 0.0031 & 0.6880 $\pm$ 0.0303 \\
 & QSLO $\uparrow$ & 0.3177 $\pm$ 0.0081 & 0.2963 $\pm$ 0.0058 & 0.2913 $\pm$ 0.0046 \\
 & VHR $\downarrow$ & 0.2777 $\pm$ 0.0049 & 0.2877 $\pm$ 0.0015 & 0.2593 $\pm$ 0.0206 \\
\midrule
\multirow{3}{*}{\textsc{Extrinsic}} & Precision $\uparrow$ & 0.7473 $\pm$ 0.0150 & 0.6673 $\pm$ 0.0125 & 0.7170 $\pm$ 0.0488 \\
 & Recall $\uparrow$ & 0.7450 $\pm$ 0.0145 & 0.6630 $\pm$ 0.0131 & 0.6903 $\pm$ 0.0318 \\
 & F1 $\uparrow$ & 0.7440 $\pm$ 0.0145 & 0.6637 $\pm$ 0.0129 & 0.6983 $\pm$ 0.0379 \\
\midrule
\multirow{1}{*}{\textsc{Stats}} & Solve Rate $\uparrow$ & 0.6937 $\pm$ 0.0202 & 0.8164 $\pm$ 0.0023 & 0.7708 $\pm$ 0.0481 \\
\bottomrule
\end{tabular}
\caption{Granite Aggregate results across $S=5,10,15$. Values are reported as mean $\pm$ sample standard deviation computed over the three iterations.}
\label{tab:granite_agg}
\end{table*}

\end{document}